# RED: Residual Estimation Diffusion for Low-Dose PET Sinogram Reconstruction

Xingyu Ai, Bin Huang, Fang Chen, Liu Shi, Binxuan Li, Shaoyu Wang, Qiegen Liu, *Senior Member, IEEE*

*Abstract*—Recent advances in diffusion models have demonstrated exceptional performance in generative tasks across various fields. In positron emission tomography (PET), the reduction in tracer dose leads to information loss in sinograms. Using diffusion models to reconstruct missing information can improve imaging quality. Traditional diffusion models effectively use Gaussian noise for image reconstructions. However, in low-dose PET reconstruction, Gaussian noise can worsen the already sparse data by introducing artifacts and inconsistencies. To address this issue, we propose a diffusion model named residual estimation diffusion (RED). From the perspective of diffusion mechanism, RED uses the residual between sinograms to replace Gaussian noise in diffusion process, respectively sets the low-dose and full-dose sinograms as the starting point and endpoint of reconstruction. This mechanism helps preserve the original information in the low-dose sinogram, thereby enhancing reconstruction reliability. From the perspective of data consistency, RED introduces a drift correction strategy to reduce accumulated prediction errors during the reverse process. Calibrating the intermediate results of reverse iterations helps maintain the data consistency and enhances the stability of reconstruction process. Experimental results show that RED effectively improves the quality of low-dose sinograms as well as the reconstruction results. The code is available at: https://github.com/yqx7150/RED.

*Index Terms*—Low-dose PET, sinogram reconstruction, diffusion model, non-Gaussian noise, drift correction.

## I. INTRODUCTION

Medical imaging techniques are essential to modern medicine in disease diagnosis [1-3]. Positron emission tomography (PET) is one of these imaging techniques, which reveals changes in metabolic activity and biochemical processes by detecting tracers that are injected into the body. Tracers such as FDG accumulate in organs and lesions over time. Due to its radioactivity, reducing the injection dose helps mitigate the harm to human body. However, it introduces additional noise and artifacts to the imaging results, which impact diagnostic accuracy [4] [5]. When these tracers are annihilated by electrons, pairs of gamma photons are produced and subsequently detected by PET scanner to generate sinograms.

Sinograms can be transferred to image domain via back-projected algorithms. Due to the presence of random noise in raw data, PET reconstruction is an ill-posed inverse problem. Numerous studies have demonstrated the capability to reconstruct images from sinograms. Filtered back projection (FBP) reconstructs images by back-projecting filtered projection data, achieving rapid image reconstruction. Still, its performance declines in high-noise conditions. In contrast, maximum likelihood expectation maximization (MLEM) optimizes the likelihood of sinograms through iterative reconstruction, but it requires substantial time to converge. Although ordered subset expectation maximization (OSEM) [6] enhances efficiency of iteration, it struggles to compensate for missing information in low-dose situation. Hence, finding effective methods to improve the quality of low-dose sinograms become a crucial area of research.

With the rapid development of machine learning, an increasing number of deep learning-based methods have been applied to PET imaging [7-10]. Cui *et al.* presents a method that integrates sinograms, images, and frequency data to perform low-dose PET reconstruction [11]. Hong *et al.* developed a data-driven sinogram method for single-image super-resolution based on a novel deep residual convolutional neural network, enhancing its effectiveness in PET imaging [12]. William *et al.* [13] presented a sinogram repair network that effectively addresses the missing data caused by block detector failures. Meanwhile, generative models have presented strong capabilities in medical imaging. Generative adversarial networks (GANs) [14] involve generating and discriminating networks to enhance generation quality. These networks have seen extensive use in the medical imaging domain[15-18]. Nonetheless, GANs face challenges due to their unstable training process, which is prone to training failures [19].

Recently, diffusion models have emerged as the leading generative models [20-24], which reconstruct images by predicting and removing noise [25]. Denoising diffusion probabilistic model (DDPM) [26] generates images directly from noise. This method iteratively generates high-quality images, rather than producing them in a single prediction. Guan *et al.* [27] proposed the generative modeling in sinogram domain (GMSD) approach to perturb sinograms with multi-scale noise and interpolate data. Shen *et al.* [28] introduced a novel approach by adding Gaussian noise in the projection domain to simulate the formation of PET noise. Huang *et al.* [29] pre-trained an encoding network, then applied a latent diffusion model [30] to enhance the quality of latent embedding. Because DDPM requires a large number of iterative steps, denoising diffusion implicit model (DDIM)

This work was supported by National Natural Science Foundation of China (621220033, 62201193). (X. Ai and B. Huang are co-first authors) (Corresponding authors: S. Wang and Q. Liu)

X. Ai, L. Shi, S. Wang, Q. Liu and F. Chen are with School of Information Engineering, Nanchang University, Nanchang, China ({aixingyu.aiden, shiliu, wangshaoyu, liuqiegen}@ncu.edu.cn; chenfang@email.ncu.edu.cn)

B. Huang is with School of Mathematics and Computer Sciences, Nanchang University, Nanchang, China (huangbin@email.ncu.edu.cn)

B. Li is with Institute of Artificial Intelligence, Hefei Comprehensive National Science Center, Hefei, China (libingxuan@iai.ustc.edu.cn).

[31] significantly improves sampling efficiency by generalizing DDPM into a non-Markovian process. These modifications allow the reverse process to achieve similar outcomes with fewer sampling steps, and without the need to add noise.

Traditional diffusion models erode input images by adding Gaussian noise through the forward process. Although the noise mechanism helps enhance the diversity of generation results, excessive variability in PET imaging can decrease the fidelity of reconstruction results. Furthermore, adding Gaussian noise to sinograms can produce artifacts after back-projection to image domain [32]. Artifacts and noise can lead to misdiagnoses, further reducing the reliability of the results. This effect is more pronounced in low-dose sinograms, where the added artificial noise makes already incomplete information even harder to discern.

To address these aforementioned issues, this study proposes a diffusion mechanism called residual estimation diffusion (RED), which adjusts the diffusion process to be more suitable for low-dose PET sinogram reconstruction. RED constructs the diffusion process by minimizing the residual between low-dose and full-dose sinograms. Instead of adding Gaussian noise to data, RED directly uses low-dose sinogram as the starting point and full-dose sinogram as the endpoint of reverse process. In this case, the low-dose sinogram can be considered as an already eroded version of the full-dose one, and no Gaussian noise needs to be added. Since RED avoids compromising the quality of the input in any form, the reverse process is a deterministic sampling process. Although this process reduces the uncertainty of the generation results, the accumulation of prediction errors can lead to data drift. This phenomenon is caused by inaccurate predictions and is more significant under non-Gaussian noise conditions. Therefore, a cascaded estimation module (CEM) is designed for the reverse process. CEM first estimates the residuals from current sinogram, and then corrects the accumulated errors of it. When sampling training data, instead of directly using actual residual between sinograms, RED uses the predicted residuals to generate intermediate samples. This strategy simulates the occurrence of drift, improves data consistency during training and reconstruction. The main contributions of this paper can be summarized as follows:

- **Residual Estimation Diffusion Mechanism.** Compared to traditional diffusion models, RED replaces Gaussian noise with the residuals between low-dose and full-dose sinograms and reconstructs sinograms via minimizing the residuals. By setting the low-dose and full-dose sinograms as the starting and ending points of reconstruction, the low-dose sinograms can be viewed as the results of adding residuals to the full-dose sinograms and do not need further degradation. This approach avoids extra Gaussian noise, thereby enhancing the reliability of results.
- **Drift Correction Strategy.** To mitigate the impact of data drift on reconstruction results, this study introduces a drift correction mechanism in the reverse process. During reconstruction, CEM estimates the residual and then calibrates the intermediate sinograms. It is trained in a staged manner. CEM first train a network to predict the accurate residuals. After that, the pre-trained network is used to make a single-step prediction to obtain drifted samples. These samples help reproduce the intermediate results of reconstruction, enhancing data consistency and stability in the reverse process.

The rest of the manuscript is organized as follows: Section II introduces low-dose PET sinograms as well as the background knowledge of diffusion models. Section III presents the internal details of RED. Experimental results are revealed in Section IV, followed by additional discussions in Section V, with the final conclusion drawn in Section VI.

## II. PRELIMINARY

### A. PET Sinogram Reconstruction

The PET scanner organizes acquired raw data into independent parallel slices. Each slice includes a list of coincidence events. Event connects the two detectors along a line of response (LOR), indicating the path of positron emissions. The raw data are structured in a sinogram, and the reconstruction process can be framed as an inverse problem:

$$b = Ax + n, \qquad (1)$$

where $x$ is the radioactive activity map, $b$ is the reconstructed image and $n$ represents noise. $A$ is the system matrix that encapsulates the tomographic geometry and physical factors of the imaging system.

Various reconstruction algorithms can be applied to process sinograms. MLEM reconstructs sinogram by maximizing the likelihood function of observed data through forward and backward projection. OSEM further improves reconstruction efficiency by dividing the entire sinogram dataset into subsets.

In low-dose PET scenarios, a reduced tracer dose leads to fewer photons reaching the detector, compromising imaging accuracy [33]. Fig. 1 illustrates the sinograms and images reconstructed by OSEM. Traditional reconstruction algorithms struggle to reconstruct low-dose sinograms due to the substantial information loss, while deep learning-based generative models offer new approaches to address these challenges.

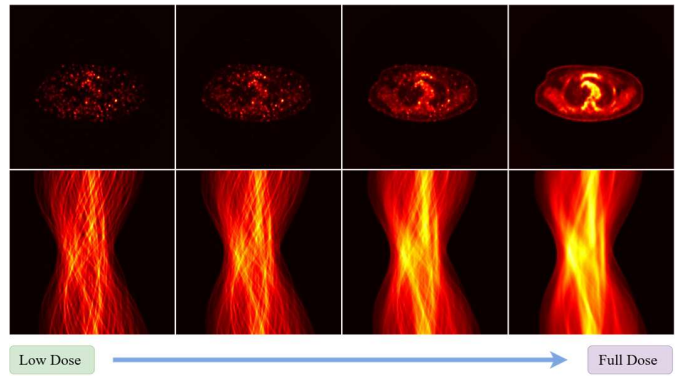

**Fig. 1.** The low-dose data (left) has worse quality compared to the full-dose data (right). Graininess and discontinuities increase as the lines in the sinogram become sparser in the projection domain.

### B. Noise Removal Diffusion Mechanism

Diffusion models are generative models that iteratively reconstruct images degraded by noise. Diffusion mechanism is formed by forward and reverse processes. For the forward process, DDPM operates on the input image $x_0$ and gradually transforms it into Gaussian noise $x_t \sim \mathcal{N}(0, I)$ over $t$ iterations. Each iteration of this process is described as follows:

$$q(x_t|x_{t-1}) = \mathcal{N}(x_t; \sqrt{1-\beta_t}x_{t-1}, \beta_t I), \quad (2)$$

where $\beta_t$ is a fixed variance schedule. Since adding Gaussian noise step by step is too cumbersome, DDPM simplifies the forward sampling process by utilizing the additivity of Gaussian distributions. Let $\alpha_t$ denotes $1 - \beta_t$ and $\bar{\alpha}_t$ represents $\prod_{i=1}^{t} \alpha_i$, $x_t$ can be calculated by:

$$x_t = \sqrt{\bar{\alpha}_t} x_0 + \sqrt{1-\bar{\alpha}_t} \epsilon, \quad (3)$$

where $\epsilon$ is a noise sampled from $\mathcal{N}(0, I)$. Diffusion model generates $x_{t-1}$ using current time step data $x_t$ in the reverse process. A neural network with parameters $\theta$ is used to approximate this process:

$$p_\theta(x_{t-1}|x_t) = \mathcal{N}(x_{t-1}; \mu_\theta(x_t, t), \Sigma_\theta(x_t, t)). \quad (4)$$

Due to the Markov property of the diffusion process, the denoising process also requires generating $x_{t-1}$ from $x_t$ at each step. As a result, DDIM proposes a non-Markovian sampling process by redefining the sample generation method. In this context, the reverse process does not need to denoise step by step, and $x_t$ can be used to compute any $x_s$ for $s < t$. Because of that, by using $\epsilon_\theta(x_t, t)$ to predict $\epsilon$, $x_s$ can be calculated by:

$$\hat{x}_0 = x_t - \sqrt{1-\bar{\alpha}_t}\epsilon_\theta(x_t, t)/\sqrt{\bar{\alpha}_t}, \quad (5)$$

$$x_{s-1} = \sqrt{\bar{\alpha}_s}\hat{x}_0 + \sqrt{1-\bar{\alpha}_s}\epsilon_\theta(x_s, t), \quad (6)$$

where $\hat{x}_0$ denotes the current prediction of the original $x_0$. Eq. (6) allows DDIM to achieve comparable results with fewer sampling steps, which in turn improves the efficiency of reverse process.

## III. METHOD

### A. Motivation

The basic idea of RED is to directly utilize low-dose sinograms as the original input to the diffusion model, thereby avoiding compromising the integrity of the original data. Traditional diffusion models transform different data into an approximate prior distribution through Gaussian noise. This mechanism improves the generalization ability, yet adding Gaussian noise to sinograms introduces artifacts into the imaging result. Since low-dose PET sinograms suffer from missing information, extra noise may make existing features harder to distinguish, further complicating the reconstruction. Meanwhile, due to the inherent complexity of PET imaging, the types of noise present in the data may differ from Gaussian noise. Considering these factors, finding a suitable method to model the difference between low-dose and full-dose sinograms can produce a more effective reconstruction process.

Recent studies have validated the feasibility of constructing diffusion models without Gaussian noise. Cold diffusion [34] employs various types of non-Gaussian noise, such as blurring and masking, as degradation operators to process input images. It achieves effective reconstruction across various noise types, yet applying degradation operations will damage the original data. If the residual between two sinograms is used to replace the noise, a given low-dose sinogram can be viewed as the result of a full-dose sinogram with noise already added. In this situation, the low-dose sinogram can be directly used as input without the need of further degradation. For that reason, RED uses the residual instead of Gaussian noise, and iteratively minimizes the gap between sinograms during the reverse process.

Under optimal conditions, the residuals predicted by RED would precisely match the actual values. However, it is challenging for deep neural networks to achieve perfect predictions in practice [35]. Generating content that does not exist in the original data may lead to misdiagnoses, reducing the credibility of predictions in clinical scenarios. Fig. 2 illustrates the data drift in the reverse process. Accumulated errors during reconstruction may cause intermediate results to deviate from training data, which consequently reduces the consistency of data. To mitigate this, RED introduces a drift correction strategy to calibrate the sinogram, thereby reducing sample deviations throughout iterations for a more stable reverse process.

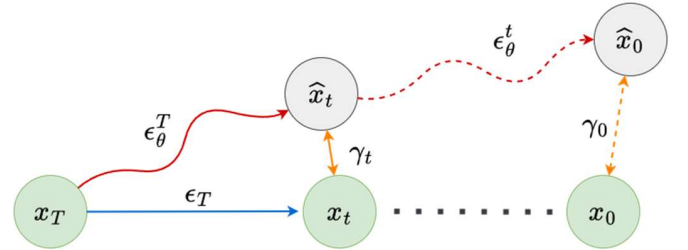

**Fig. 2.** Illustration of data drift in the reverse process. Drift $\gamma_t$ accumulates due to the imperfect prediction of residual, leading the final reconstruction result to deviate from expectations.

### B. Residual Estimation Mechanism

RED utilizes the residual term $\epsilon$ to represent the difference between low-dose sinogram $x_L$ and full-dose sinogram $x_F$ which can be described as:

$$x_L = x_F + \epsilon. \quad (7)$$

As a result, $x_L$ can be obtained from $x_F$ by adding the residual $\epsilon$ through $T$ iterations. Due to the fact that RED does not rely on Gaussian noise assumption, the forward process can be simplified to adding the residual to the full-dose image. For a given time step $t$, $x_t$ can be obtained by adding a certain proportion of residual:

$$x_t = x_F + \alpha_t \epsilon, \forall t \in \{1, \dots, T\}, \quad (8)$$

where $T$ and $\alpha_1, \cdots, \alpha_T \in [0,1)$ represent the number of diffusion steps and the residual schedule across diffusion steps. In the reverse process, $x_F$ is defined as endpoint $x_0$, and $x_L$ is defined as starting point $x_T$. When $t$ reaches the endpoint, $\alpha_0$ is set to 0, indicating that $x_L$ has completely transformed into a full-dose sinogram.

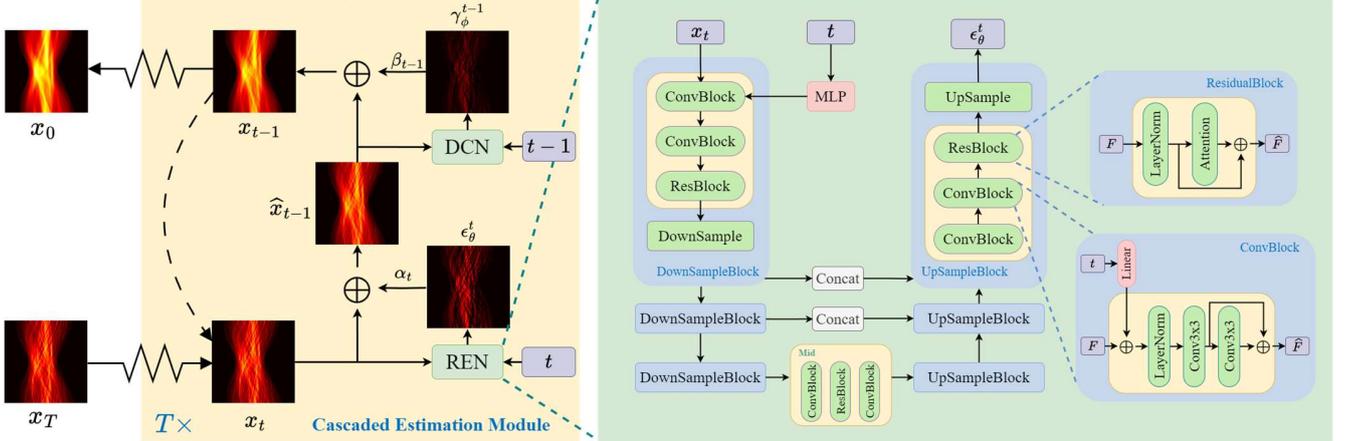

**Fig. 3.** The overall structure of RED. During the reverse process, CEM first estimates the residual to calculate $\hat{x}_{t-1}$, then corrects it to prevent excessive drift. Both REN and DCN use the same network architecture.

Based on the forward process, RED correspondingly adjusts the reverse process. Hence sample $x_{t-1}$ can be obtained by:

$$x_{t-1} = x_0 + \alpha_{t-1}\epsilon. \quad (9)$$

By substituting $x_0$ using Eq. (8), $x_{t-1}$ can be computed from $x_t$ as follows:

$$\begin{aligned} x_{t-1} &= (x_t - \alpha_t \epsilon) + \alpha_{t-1}\epsilon \\ &= x_t - (\alpha_t - \alpha_{t-1})\epsilon. \end{aligned} \quad (10)$$

This formula iteratively computes $x_{t-1}$ from $x_t$, forming the reverse process. In this case, the intermediate results can be iteratively computed through the reverse process:

$$x_t = x_T - \sum_{i=T}^{t}(\alpha_i - \alpha_{i-1})\epsilon. \quad (11)$$

Meanwhile, the reverse process is deterministic and does not require degradation. Modeling the residual between the data simplifies the diffusion process and establishes a transformation from low-dose to full-dose sinograms.

### C. Drift Correction Strategy

Due to the value of $\epsilon$ is unknown in practice, RED employs a pair of cascaded networks to estimate the residual. Fig. 3 illustrates the structure of CEM. It combined by a residual estimation network (REN) with parameters $\theta$ and another drift correction network (DCN) with parameter $\phi$. The prediction of REN can be represented as:

$$\epsilon_\theta^t = REN_\theta(x_t, t), \quad (12)$$

where $\epsilon_\theta^t$ denotes the predicted residual at time step $t$. Thus, the predicted intermediate result $\hat{x}_t$ can be denote as:

$$\hat{x}_t = x_T - \sum_{i=T}^{t}(\alpha_i - \alpha_{i-1})\epsilon_\theta^i. \quad (13)$$

The reverse process of RED divides reconstruction into multiple iterations. As a result, prediction errors accumulate as the number of iterations increases. Denoting $\delta_i$ as the predicted error at time step $i$, the actual residual $\epsilon$ can be represented as:

$$\epsilon = \epsilon_\theta^i - \delta_i. \quad (14)$$

Calibrating the errors that arise during the iterative process helps improve the data consistency. Therefore, correction can be conducted by replacing $\epsilon$ in Eq. (11):

$$\begin{aligned} x_t &= x_T - \sum_{i=T}^{t}(\alpha_i - \alpha_{i-1})(\epsilon_\theta^i - \delta_i) \\ &= \hat{x}_t + \sum_{i=T}^{t}(\alpha_i - \alpha_{i-1})\delta_i, \end{aligned} \quad (15)$$

where $\hat{x}_t$ means the reconstruction result without drift correction. Letting $\gamma_t = \sum_{i=T}^{t}(\alpha_i - \alpha_{i-1})\delta_i$, Eq. (15) can be simplified as:

$$\gamma_t = x_t - \hat{x}_t. \quad (16)$$

This formula indicates that the correction of data drift can be obtained by estimating the residual between the current predicted $\hat{x}_t$ and actual $x_t$. RED uses another drift correction network (DCN) with parameter $\phi$ to predict the drift correction term $\gamma_\phi^t$:

$$\gamma_\phi^t = DCN_\phi(\hat{x}_t, t). \quad (17)$$

REN cascades with DCN to form CEM. In the reverse process, CEM first uses REN to calculate $\epsilon_\theta^t$, then feeds the predicted $\hat{x}_{t-1}$ into the DCN to obtain $\gamma_\phi^{t-1}$. Hence the reverse process can use the following equations to calculate $x_{t-1}$ for the next iteration step:

$$\hat{x}_{t-1} = x_t - (\alpha_t - \alpha_{t-1})\epsilon_\theta^t \quad (18)$$

$$x_{t-1} = \hat{x}_{t-1} + \beta_{t-1}\gamma_\phi^{t-1}, \forall t \in \{1, \dots, T\} \quad (19)$$

where $\beta_1, \cdots, \beta_T \in [0,1)$ denotes a schedule to adjust the drift correction. In summary, CEM predicts the overall residual through $x_t$ and corrects the prediction errors by constraining the drift in $\hat{x}_{t-1}$.

### D. Training and Reconstruction

RED performs reconstruction using CEM, which is composed of REN and DCN in cascade. The training of CEM is carried out in stages: first, REN is trained to estimate the residuals, and then the pre-trained REN generates drifted samples to train DCN.

*REN Training:* REN uses a modified U-Net architecture to estimate residuals based on the input sinogram and time step. To construct training samples, a low-dose sinogram $x_L$ and its corresponding full-dose sinogram $x_F$ are selected from the dataset. A time step $t$ is then sampled from a uniform distribution between 0 and the maximum time step $T$. Next, REN utilizes Eq. (8) to calculate $x_t$ as the training sample. Time step $t$ and $x_t$ are fed into REN to generate predicted residuals. During training, predicted $\epsilon_\theta^t$ is compared to the actual $\epsilon$, and the network is optimized by minimizing the mean squared error (MSE) loss:

$$\mathcal{L}_{MSE} = \mathbb{E}_{t,(x_L,x_F)}[\|\epsilon - \epsilon_\theta^t\|_2^2]. \qquad (20)$$

At the same time, the training introduces another loss function based on structural similarity (SSIM) to evaluate the restoration of the estimated residuals:

$$\mathcal{L}_{SSIM}(x,y) = 1 - \frac{(2\mu_x\mu_y + c_1)(2\sigma_{xy} + c_2)}{(\mu_x^2 + \mu_y^2 + c_1)(\sigma_x^2 + \sigma_y^2 + c_2)}, \qquad (21)$$

where x and y denote the two input images, $c_1$ and $c_2$ are two coefficients. $\mu_x$ and $\mu_y$ measure the brightness of input image. $\sigma_x$ and $\sigma_y$ are the variances related to contrast. $\sigma_{xy}$ is the covariance between images x and y used to measure the structure similarity. Thus, by minimizing these two loss functions, the total loss and objective function can be written in the following form:

$$\mathcal{L}_{REN} = \mathcal{L}_{MSE} + \mathcal{L}_{SSIM}(\hat{x}_0, x_0); \ \theta^* = arg \min_\theta \mathcal{L}_{REN}. \qquad (22)$$

where $\hat{x}_0$ is the reconstructed sinogram. By minimizing the error of prediction, the similarity between low-dose and full-dose sinograms can be enhanced.

*DCN Training:* RED utilizes DCN for data drift correction, and DCN uses a pre-trained REN to generate drifted training samples. For a random time step $t$, samples can be generated through reverse process. However, performing a complete reverse process would significantly increase the computational complexity. To improve training efficiency, DCN uses REN for a one-step prediction and uses the predicted residual to generate drifted training sample $\hat{x}_t$. REN first estimates the residual of $x_T$. The predicted $\epsilon_\theta^T$ is then interpolated with $\epsilon$ using a random coefficient $\lambda \sim Uniform(0,1)$, and the result is substituted into Eq. (8) to generate drifted training samples:

$$\hat{x}_t = x_F + (1 - \alpha_t)(\lambda \epsilon_\theta^T + (1 - \lambda)\epsilon). \qquad (23)$$

Since $x_T$ suffers more information loss compared to the intermediate results, $\epsilon_\theta^T$ can produce more pronounced drifted samples for training. When the predictions of REN become more accurate, the drift will correspondingly decrease. The deviation between the generated sample and the actual value can be calculated using Eq. (16), and the optimization objective is to minimize MSE between the predicted and the actual drift. Hence, the objective function of DCN can be written as follows:

$$\phi^* = arg \min_\phi \mathbb{E}_{t,(\hat{x}_t,x_t)}\left[\|\gamma_t - \gamma_\phi^t\|_2^2\right]. \qquad (24)$$

*RED Reconstruction:* Since the forward process of RED is not a Markov chain, accelerated sampling methods can be employed in the reverse process to speed up sampling. For a given number of iterations $T_s$, the skipped sampling steps $r$ can be obtained by calculating the ratio of $T_s$ to the original maximum time step $T_{MAX}$.

$$r = T_s/T_{MAX}. \qquad (25)$$

By modifying Eq. (18), for a given sample $x_t$, the previous sample $x_s$ where $s = t - r$ can be calculated by:

$$\hat{x}_{t-r} = x_t - (\alpha_t - \alpha_{t-r})\epsilon_\theta^t. \qquad (26)$$

In this case, $t - r$ represents the next sampling time step after skipping a certain ratio of reverse iterations. Algorithm 1 presents the specific process of training stage and reconstruction stage of RED.

---

**Algorithm 1: Residual Estimation Diffusion (RED)**

**Training Stage**

**Require**: Schedule $\alpha$ and $\beta$; Learning rate $\eta$, Maximum time step $T$; Number of steps $N$. Parameters $\theta$ and $\phi$

**Repeat**

**for** $i = 1,2,\ldots,N$ **do**
  $x_L \sim q(x_L), x_F \sim q(x_F), t \sim Uniform(0,T)$
  Sample $x_t$ according to Eq. (8)
  $\epsilon_\theta^i \leftarrow REN_\theta(x_t, t)$
  Take gradient decent step via Eq. (22)
  $\theta \leftarrow \theta - \eta\nabla_\theta \mathcal{L}_{REN}$
**end for**

**for** $i = 1,2,\ldots,N$ **do**
  $x_L \sim q(x_L), x_F \sim q(x_F), t \sim Uniform(0,T)$
  Sample $x_t$ according to Eq. (8)
  $\epsilon_\theta^i \leftarrow REN_\theta(x_L, t)$
  Sample $\hat{x}_t$ according to Eq. (23)
  $\gamma_\phi^t \leftarrow DCN_\phi(\hat{x}_t, t)$
  Take gradient decent step via Eq. (24)
  $\phi \leftarrow \phi - \eta\nabla_\phi \mathbb{E}_{t,(\hat{x}_t,x_t)}\left[\|\gamma_t - \gamma_\phi^t\|_2^2\right]$
**end for**

**Until converged**

**Reconstruction Stage**

**Require**: Schedule $\alpha$ and $\beta$; Maximum time step $T_{Max}$; Number of iterations $T_s$; Trained parameters $\theta^*$ and $\phi^*$
$x_L \sim q(x_L)$
$r = T_s/T_{MAX}, t = T_{MAX}, x_t = x_L$
**for** $s = T_s,\ldots,1,0$ **do**
  $\hat{x}_{t-r} = x_t - (\alpha_t - \alpha_{t-r})REN_{\theta^*}(x_t, t)$
  $x_{t-r} = \hat{x}_{t-r} - \beta_{t-r}DCN_{\phi^*}(\hat{x}_{t-r}, t-r)$
  $t = t - r$
**end for**
**Return** $x_0$

---

This training process can be iterated multiple times to enhance the generalization ability and improves the final reconstruction performance.

## IV. EXPERIMENTS

To validate the performance of RED, this section compares it with OSEM [6], U-Net [36], DDIM [31], and cold diffusion (CD) [34] in reconstruction tasks. The raw data used in the dataset was obtained through OSEM, while other methods first take low-dose sinograms as input and then utilize the FBP algorithm to reconstruct image results [37]. The experiment introduces DDIM and CD as different types of diffusion models to compare their effectiveness with RED. Additionally, U-Net was included in the comparison to assess the performance of RED with supervised methods to provide more comprehensive comparisons. All models were trained using the same device and dataset to ensure consistency and fairness.

*UDPET:* RED is trained via public dataset from MICCAI 2022 Ultralow-Dose PET (UDPET) Imaging Challenge. Each PET image corresponds to a low-dose PET with a specific dose reduction factor (DRF) and was reconstructed from counts of a time window resampled at the midpoint of acquisition. Each patient includes imaging results under different doses and each dose comprising 673 slices. This experiment utilized 18 F-FDG PET imaging subjects, acquired using uEXPLORER under DRFs of 4, 20, and 100, with each case comprising 500 patients, resulting in a total of 336500 2D slices for training purposes. Additionally, 1,000 extra slices were selected separately to validate the effectiveness at different dose levels. The images in the dataset were reconstructed using OSEM, and the corresponding sinograms were obtained by back-projecting the images [38]. Sinograms were slightly resized from $360 \times 360$ to $352 \times 352$. All the training data were normalized before being input into the network. The weight and injection dose of patient is used to regularize the image, and these parameters are recorded to restore predictions to the original data range.

*Patient Data:* To further assess the generalization capability of these models, experiments selected data from 10 patients sampled at 25% dose from DigitMI 930 PET/CT scanner for validation. This scanner was developed by RAYSOLUTION Healthcare Co., Ltd. and features All-Digital PET detectors with an axial field-of-view (AFOV) of 30.6 cm. Each patients covered 4 to 8 beds, with scan times ranging from 45 seconds to 3 minutes per bed. Low-dose PET data was obtained by resampling the listmode data into 2 ms intervals, retaining 1 ms of data per cycle while discarding the rest.

*Experimental Setup:* RED was trained on a GPU (NVIDIA GeForce RTX 3090 24GB) using the PyTorch framework. The AdamW optimizer [39] was used, with an initial learning rate set to $1 \times 10^{-4}$, which was gradually decayed by half every $1 \times 10^5$ iterations. The maximum sampling step $T_{MAX}$ was set to 500 and $T_s$ was set to 30. Various evaluation metrics were employed in the experiment: peak signal-to-noise ratio (PSNR), structural similarity index (SSIM), and normalized root mean squared error (NRMSE).

### A. Reconstruction Experiments

*Imaging Result Comparisons:* The evaluation results are presented in Table I, with RED achieving the best performance across all evaluation metrics. Due to the high quality of the initial data, all models perform relatively well under DRF 4, with RED achieving the highest PSNR value of 39.57 dB. Furthermore, RED consistently delivered the best results under DRF 20. This trend became even more pronounced at DRF 100, where RED outperforms OSEM by 8.08 dB and surpasses the second-ranked CD by 2.23 dB in terms of PSNR. Figs. 4-6 display the sinograms and reconstruction results at different dose levels. Overall, the results of RED are smoother while retaining more details. At DRF 4, images of RED exhibit fewer noise and artifacts, whereas other methods signify more graininess within the organs. Under DRF 20, the data quality degraded, and the unsupervised diffusion methods perform poorly. Although U-Net manages to reconstruct the lost information, it produces blurry images. In contrast, results of RED are more similar to the full-dose images, achieving smooth interiors while preserving sharp edges. At DRF 100, the result of OSEM becomes barely recognizable, and U-Net introduces considerable blurring. Both DDIM and CD recover few details. Despite the extremely poor conditions, RED still manages to restore part of the information to maintain structural integrity.

TABLE I
AVERAGE QUANTITATIVE RECONSTRUCTION QUALITY OF DIFFERENT METHODS AT DRF 4, 20, 100.

| Method | DRF 4 (25%) | | | DRF 20 (5%) | | | DRF 100 (1%) | | |
|---|---|---|---|---|---|---|---|---|---|
| | PSNR↑ | SSIM↑ | NRMSE↓ | PSNR↑ | SSIM↑ | NRMSE↓ | PSNR↑ | SSIM↑ | NRMSE↓ |
| OSEM [6] | 36.82 | 0.956 | 0.037 | 29.48 | 0.896 | 0.063 | 20.13 | 0.802 | 0.109 |
| U-Net [36] | 37.58 | 0.967 | 0.015 | 32.93 | 0.929 | 0.023 | 24.30 | 0.844 | 0.042 |
| DDIM [31] | 36.95 | 0.937 | 0.024 | 31.19 | 0.906 | 0.022 | 23.89 | 0.839 | 0.046 |
| CD [34] | 38.21 | 0.972 | 0.022 | 33.21 | 0.937 | 0.020 | 25.98 | 0.851 | 0.031 |
| **RED** | **39.57** | **0.977** | **0.012** | **34.93** | **0.952** | **0.018** | **28.21** | **0.877** | **0.028** |

The comparison of reconstructed sinograms from different methods provides more intuitive evaluation. Table II directly presents the evaluation metrics in the projection domain under DRF 20. RED achieved the best results across all evaluation metrics and demonstrated excellent capability with a PSNR improvement of 5.04 dB compared to OSEM. While other methods indicated lower similarity, which may cause artifacts and noise after back-projection.

TABLE II
AVERAGE QUANTITATIVE QUALITY OF SINOGRAM AT DRF 20

| Method | DRF 20 (5%) | | |
|---|---|---|---|
| | PSNR↑ | SSIM↑ | NRMSE↓ |
| OSEM [6] | 30.67 | 0.824 | 0.072 |
| U-Net [36] | 32.54 | 0.837 | 0.035 |
| DDIM [31] | 30.51 | 0.863 | 0.039 |
| CD [34] | 31.42 | 0.916 | 0.029 |
| **RED** | **35.71** | **0.947** | **0.027** |

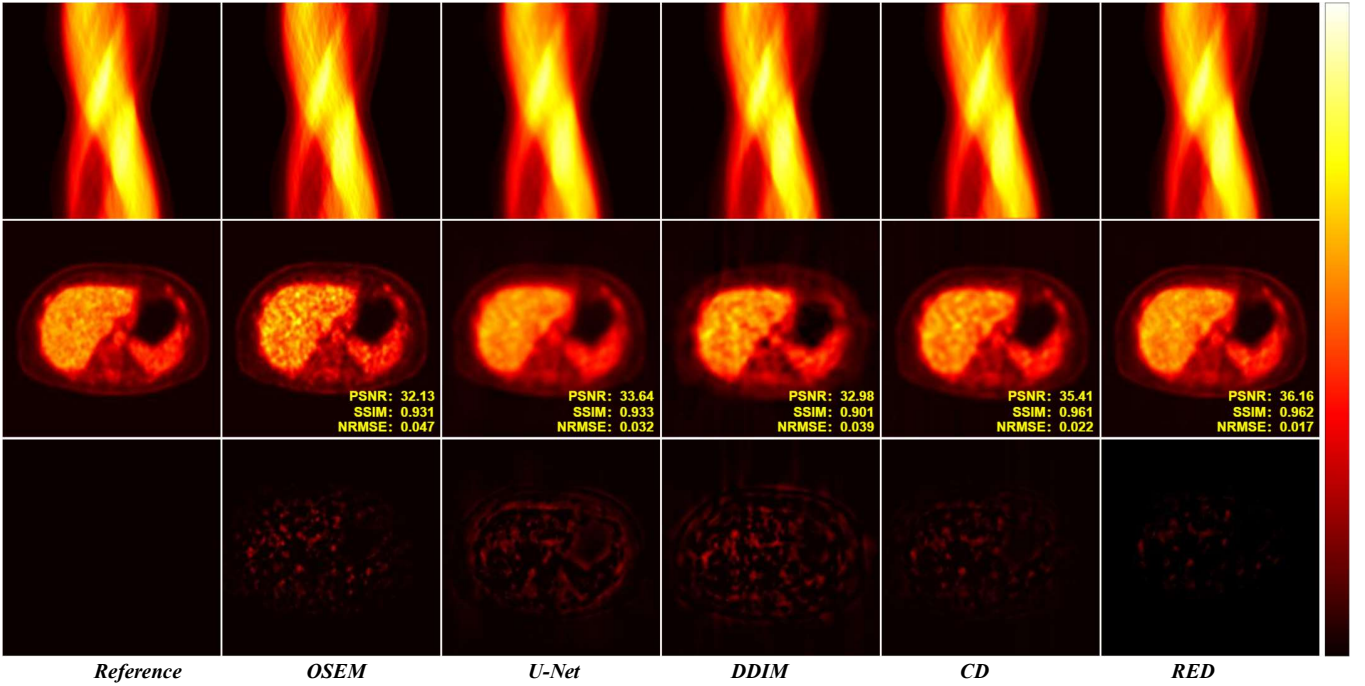

Fig. 4. Comparison of reconstruction under **DRF 4** using different methods. The first row illustrates the sinograms, the second row presents the reconstruction results, and the last row displays the residuals.

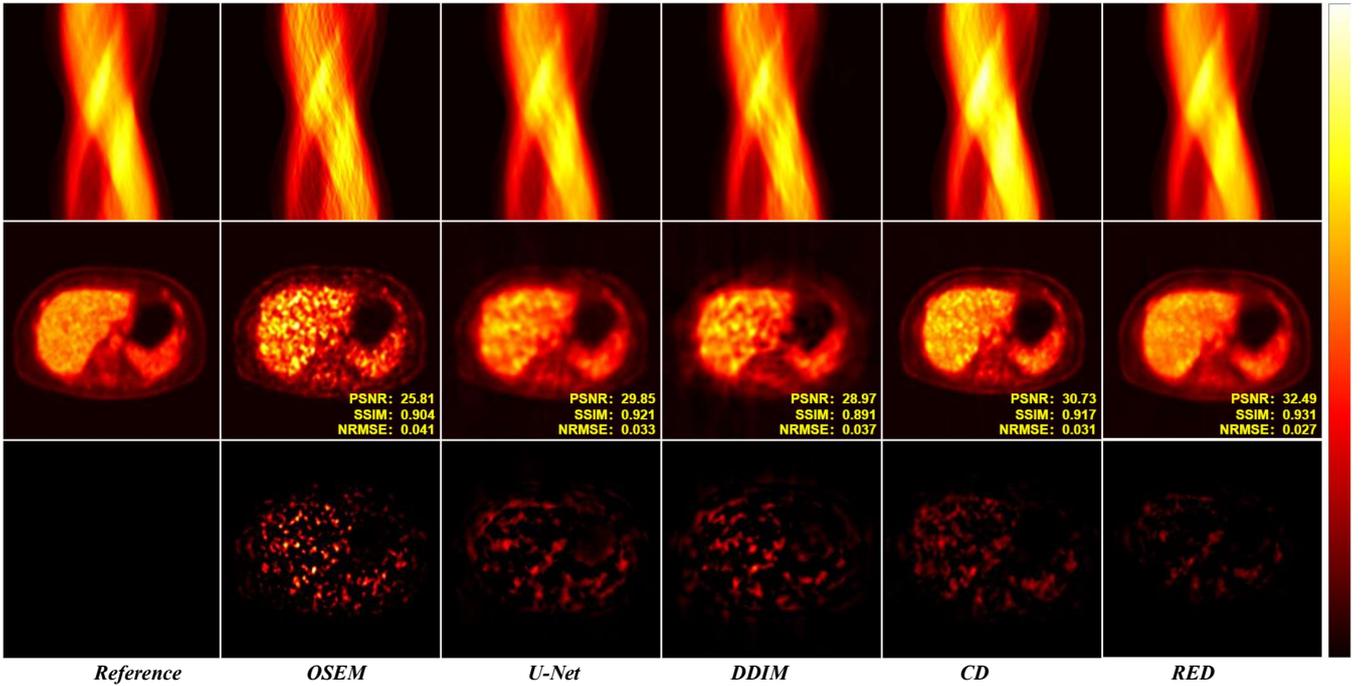

Fig. 5. Comparison of reconstruction under **DRF 20** using different methods. The first row illustrates the sinograms, the second row presents the reconstruction results, and the last row displays the residuals.

*Lesion Comparisons:* PET imaging is widely used for disease diagnosis in the brain [40] and lung [41]. For this experiment, slices 80 to 450 from each patient were selected as the critical area for comparison. This range includes brain and torso, excluding the leg areas and blank scanning slices. The evaluation of critical area under DRF 20 is presented in Table III. RED achieved the best performance and demonstrated its effectiveness of imaging critical areas.

TABLE III
AVERAGE QUANTITATIVE QUALITY AT CRITICAL AREA.

| Method | DRF 20 (5%) | | |
|---|---|---|---|
| | PSNR↑ | SSIM↑ | NRMSE↓ |
| OSEM [6] | 28.97 | 0.883 | 0.034 |
| U-Net [36] | 31.18 | 0.904 | 0.029 |
| DDIM [31] | 30.22 | 0.891 | 0.032 |
| CD [34] | 31.42 | 0.916 | 0.029 |
| **RED** | **32.78** | **0.957** | **0.026** |

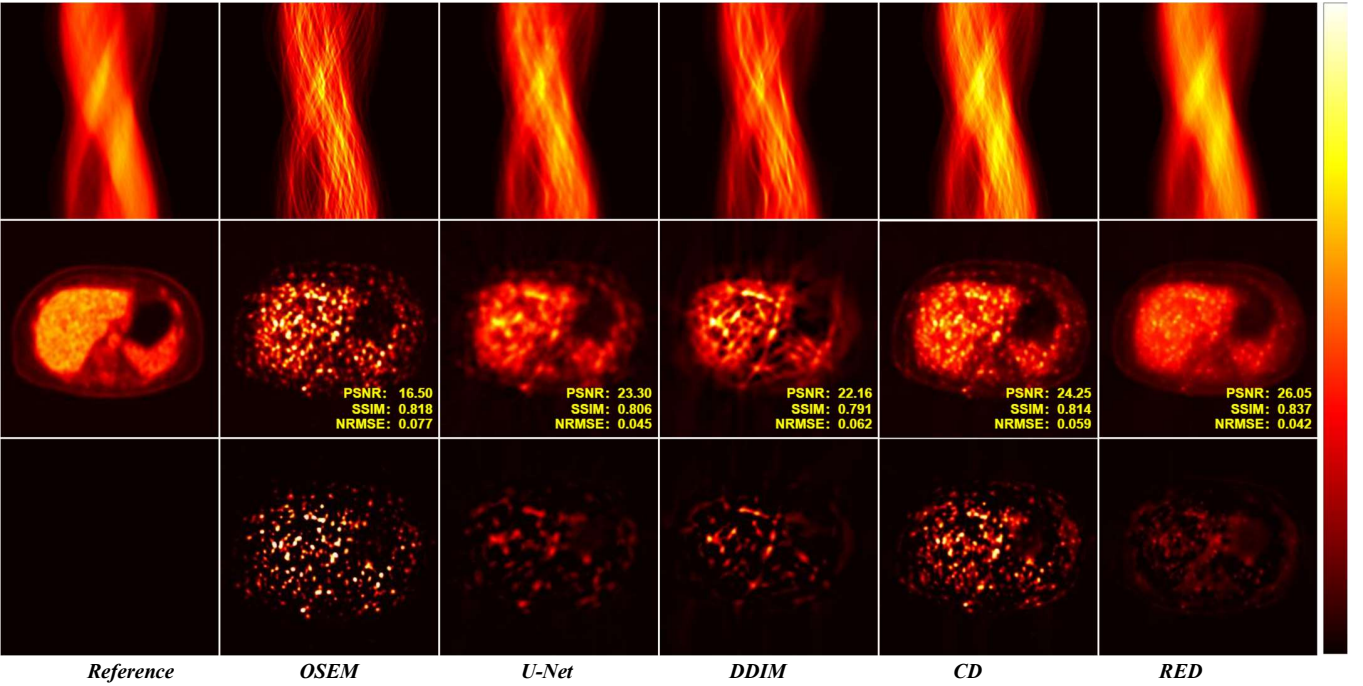

**Fig. 6.** Comparison of reconstruction under **DRF 100** using different methods. The first row illustrates the sinograms, the second row presents the reconstruction results, and the last row displays the residuals.

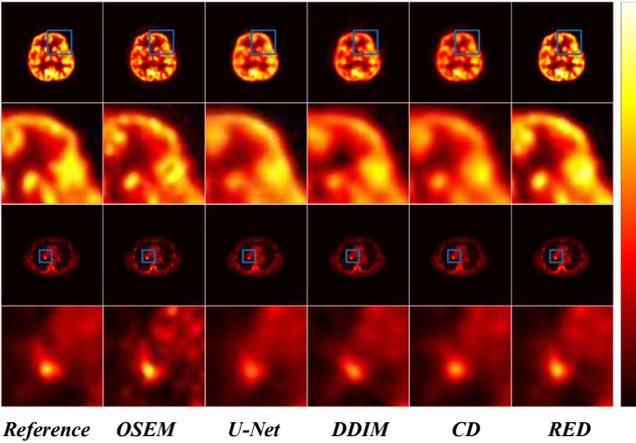

**Fig. 7.** The illustration of the brain and lung regions under DRF 20. RED preserves more details while also producing smoother results.

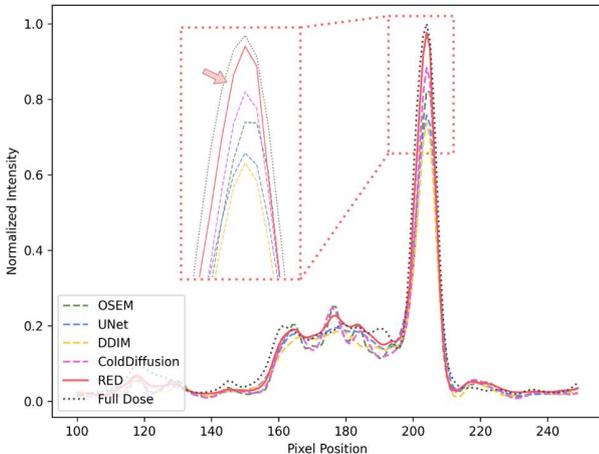

**Fig. 8.** The profile lines of each model after normalization show that the recovery values of RED are closest to the real data.

Furthermore, Fig. 7 provides a more intuitive comparison of the imaging results for the brain and lung regions. Under this circumstance, RED not only preserved image details but also avoided excessive smoothing. The profile of the lesion area is expressed in Fig. 8. The imaging results of RED are closest to the true values. This experiment proves the reliability of RED, whereas other methods exhibited more noticeable discrepancies

### B. Ablation Study

RED employs DCN to correct the iterative process during reconstruction and introduces an SSIM-based loss function when training the REN. To evaluate the contribution of these strategies, Table IV presents the changes in evaluation metrics under DRF 20 after removing the drift correction (DC) and the SSIM-based loss (SL). The results signify that removing either of these measures led to a decline in performance, whereas the RED model, which simultaneously utilized both methods, achieved the best test results, demonstrating the effectiveness of both data correction and the reconstruction loss function.

TABLE IV
IMPACT OF DIFFERENT COMPONENTS AT DRF 20.

| Metrics | (w/o) DC+ SL | (w/o) DC | (w/o) SL | RED |
|---|---|---|---|---|
| PSNR↑ | 32.12 | 33.91 | 33.72 | **34.93** |
| SSIM↑ | 0.914 | 0.933 | 0.947 | **0.952** |
| NRMSE↓ | 0.035 | 0.021 | 0.023 | **0.018** |

### C. Generalization Analysis

To further validate the generalization performance of RED, the experiment employed the Patient Dataset for performance evaluation. Fig. 9 manifests a comparison of reconstruction results of different models on this dataset, while Table V lists the performance metrics for each model. The experimental results present that RED has strong generalization capability, maintaining good performance even when the dataset is changed.

TABLE V
AVERAGE QUANTITATIVE QUALITY ON PATIENT DATASET.

| Method | DRF 20 (5%) | | |
|---|---|---|---|
| | PSNR↑ | SSIM↑ | NRMSE↓ |
| OSEM [6] | 27.39 | 0.875 | 0.032 |
| U-Net [36] | 29.08 | 0.898 | 0.034 |
| DDIM [31] | 27.57 | 0.877 | 0.041 |
| CD [34] | 28.02 | 0.913 | 0.039 |
| **RED** | **31.78** | **0.931** | **0.027** |

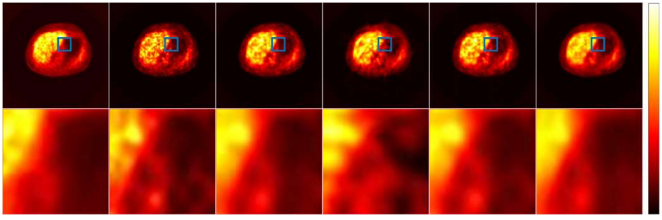

*Reference*  *OSEM*  *U-Net*  *DDIM*  *CD*  *RED*

**Fig. 9.** Comparison of the reconstruction results under **DRF 20** using different methods. The first row shows the sinograms used for reconstruction, and the second row is the reconstruction results.

### D. Experiment of Reverse Process

The diffusion models compared in the above experiments involve different diffusion strategies: DDIM reconstructs images from Gaussian noise, and the deblurring CD generates clear images from blurred ones. Both methods erode low-dose images during processing, whereas RED directly reconstructs from the low-dose sinogram. During the diffusion process, DDIM reconstructs sinogram with added noise and spends most of its iterative steps for noise removal. Meanwhile, CD retains initial information, leading to a higher initial performance.

The reverse process for these models is presented in Fig. 10. As the diffusion process continues, CD exhibited excessive deblurring which ultimately decreased the imaging quality. In contrast, since RED retains the original information from the low-dose sinogram, it starts with a high initial similarity and maintains continuous quality improvement throughout the reverse process, achieving the best reconstruction results.

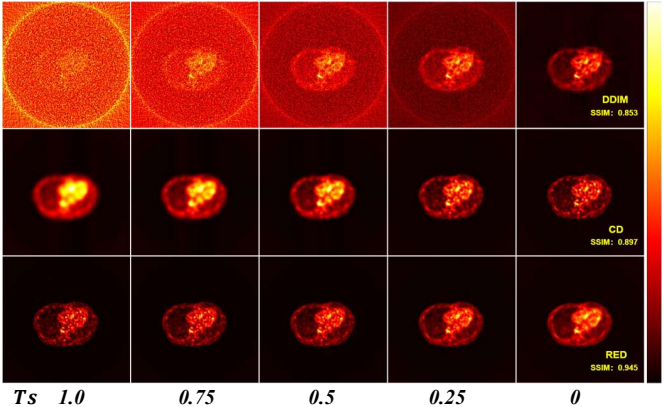

*Ts*  *1.0*  *0.75*  *0.5*  *0.25*  *0*

**Fig. 10.** Comparison of the reverse process under **DRF 20** using different diffusion mechanisms.

## V. DISCUSSION

RED demonstrates excellent performance in sinogram reconstruction for low-dose PET. Avoiding the addition of noise and preserving as much of the original information as possible helps alleviate the learning difficulty. Although RED dose not further damaging the original image, experiments have suggested that RED remains effective even when Gaussian noise is added.

Table VI demonstrates the evaluation metrics under DRF 20 after adding Gaussian noise. In this situation, the residual can be defined as the difference between two images with added noise for supervised learning. Alternatively, it can be treated as pure noise, enabling the construction of RED in an unsupervised manner. This characteristic makes the method promising for use in data-limited situations or other specific environments.

TABLE VI
EFFECTIVE COMPARISONS OF MIXED GAUSSIAN NOISE AT DRF 20.

| Metrics | DDIM | Supervised RED | Unsupervised RED | RED |
|---|---|---|---|---|
| **PSNR** | 31.19 | 32.79 | 32.13 | **34.93** |
| **SSIM** | 0.906 | 0.914 | 0.902 | **0.952** |
| **NRMSE** | 0.022 | 0.021 | 0.024 | **0.018** |

Fig. 11 presents the sinogram during the reverse process and the corresponding back-projection results. RED is still able to effectively remove noise even after adding Gaussian noise, and the overall process remains fully functional. This indicates that RED offers a variety of options in selecting degradation operators such as blurring and masking, allowing it to be combined with different types of noise.

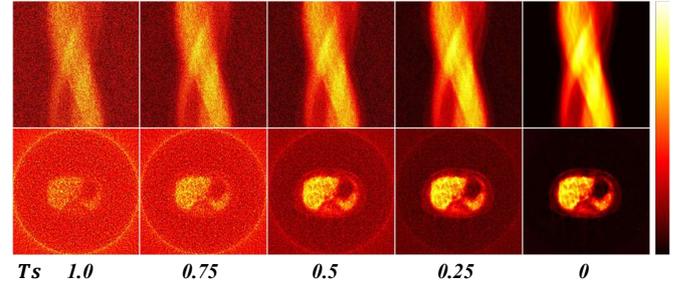

*Ts*  *1.0*  *0.75*  *0.5*  *0.25*  *0*

**Fig. 11.** The intermediate results of RED after adding Gaussian noise to inputs. RED effectively removes the added noise during the iterative process.

## VI. CONCLUSION

This study introduced RED as a high-fidelity diffusion mechanism that minimized the residuals between sinograms. Unlike standard diffusion models, RED did not introduce extra noise to low-dose sinograms, thereby preserving the integrity of the original data. To reduce the impact of accumulated prediction error during the iterations, RED introduced a drift correction mechanism and employed estimated samples during training to enhance data consistency. Further exploration of combining RED with Gaussian noise demonstrated great potential for broader applications in other domains. This offered promising directions for future research to expand the application of RED in various medical imaging tasks and beyond.